\documentclass[compsoc,conference,a4paper,10pt,times]{IEEEtran}

\AtBeginDocument{%
  \providecommand\BibTeX{{%
    \normalfont B\kern-0.5em{\scshape i\kern-0.25em b}\kern-0.8em\TeX}}}

\usepackage[linesnumbered,ruled,vlined]{algorithm2e}
\usepackage{amsmath,amssymb,amsfonts}
\usepackage{textcomp}
\usepackage{xcolor}
\usepackage{caption}
\usepackage{subcaption}
\usepackage{multirow}
\usepackage{graphicx}
\usepackage{booktabs}
\usepackage{makecell}
\usepackage{enumitem}
\usepackage{listings}
\usepackage{breakcites}
\usepackage{tcolorbox}
\usepackage{multicol} 
\usepackage{subcaption}
\usepackage{ulem}
\usepackage{wrapfig}
\usepackage[noend]{algpseudocode}
\usepackage{amsmath}
\usepackage{cite}
\usepackage{hyperref}

\begin{document}

\newcommand{\todo}[1]{\textbf{\color{red}{Ravishka: #1} }}

\newcommand{\wei}[1]{\textcolor{blue}{Wei: [#1]}}

\newcommand{\ww}[1]{\textcolor{blue}{Weihang: [#1]}}

\newcommand{\td}[1]{\textcolor{blue}{ToDo: [#1]}}

\newcommand{\zijie}[1]{{\color{purple}[Zijie: #1]}}

\newcommand{\zeqing}[1]{{\textcolor{blue}[Zeqing: #1]}}

\newcommand{\jiajun}[1]{{\color{cyan}[Jiang: #1]}}

\newcommand{\distance}{4pt}
\setlength{\textfloatsep}{\distance}

\newcommand\mycommfont[1]{\small\ttfamily\textcolor{violet}{#1}}
\SetCommentSty{mycommfont}

\lstdefinestyle{Cpp}{ 
	language=C++,
	basicstyle=\scriptsize\ttfamily, 
	breakatwhitespace=false, 
	breaklines=true, 
	captionpos=b, 
	commentstyle=\color[rgb]{0.0, 0.5, 0.69},
	deletekeywords={}, 
	escapeinside={<@}{@>},
	firstnumber=1, 
	frame=lines, 
	frameround=tttt, 
	keywordstyle={[1]\color{blue!90!black}},
	keywordstyle={[3]\color{red!80!orange}},
	morekeywords={String,int}, 
	numbers=left, 
	numbersep=-8pt, 
	numberstyle=\tiny\color[rgb]{0.1,0.1,0.1}, 
	rulecolor=\color{black}, 
	showstringspaces=false, 
	showtabs=false, 
	stepnumber=1, 
	stringstyle=\color[rgb]{0.58,0,0.82},
	tabsize=2, 
	backgroundcolor=\color{white}
}











\title{Exploiting Efficiency Vulnerabilities in Dynamic Deep Learning Systems}
\makeatletter 
\newcommand{\linebreakand}{%
  \end{@IEEEauthorhalign}
  \hfill\mbox{}\par
  \mbox{}\hfill\begin{@IEEEauthorhalign}
}
\makeatother 

\author{
\IEEEauthorblockN{Ravishka Rathnasuriya}
\IEEEauthorblockA{University of Texas at Dallas\\
USA\\
ravishka.rathnasuriya@utdallas.edu}

\and



\IEEEauthorblockN{Wei Yang}
\IEEEauthorblockA{University of Texas at Dallas\\
USA\\
wei.yang@utdallas.edu}
}

\maketitle

\begin{abstract}

The growing deployment of deep learning models in real-world environments has intensified the need for efficient inference under strict latency and resource constraints. To meet these demands, dynamic deep learning systems (DDLSs) have emerged, offering input-adaptive computation to optimize runtime efficiency. While these systems succeed in reducing cost, their dynamic nature introduces subtle and underexplored security risks. In particular, input-dependent execution pathways create opportunities for adversaries to degrade efficiency, resulting in excessive latency, energy usage, and potential denial-of-service in time-sensitive deployments.

This work investigates the security implications of dynamic behaviors in DDLSs and reveals how current systems expose efficiency vulnerabilities exploitable by adversarial inputs. Through a survey of existing attack strategies, we identify gaps in the coverage of emerging model architectures and limitations in current defense mechanisms. Building on these insights, we propose to examine the feasibility of efficiency attacks on modern DDLSs and develop targeted defenses to preserve robustness under adversarial conditions. 

\end{abstract}

\section{Introduction}

Deep learning systems have achieved state-of-the-art performance across a range of domains, including vision, language, and multimodal tasks~\cite{han2021dynamic}. However, the computational cost of deploying these models at scale remains a significant barrier, particularly in environments with constrained resources or strict latency requirements. In response, recent advances have led to the development of dynamic deep learning systems (DDLSs)~\cite{han2021dynamic}, which adjust their inference-time computation based on input complexity. By adjusting execution depth, output length, or intermediate processing stages, these systems optimize efficiency while maintaining predictive accuracy.

While DDLSs offer clear benefits in terms of performance and scalability, their adaptive nature introduces new and insufficiently understood security risks. Unlike static models~\cite{resnet16cvpr}, the computation performed by a DDLS is inherently input-dependent. This input-adaptivity, though designed to improve efficiency, also creates an opportunity for adversaries to induce worst-case behavior through carefully crafted inputs~\cite{rathnasuriya2025efficiency}. Crucially, such attacks can degrade system efficiency by manifesting as increased latency, energy consumption, or memory usage without affecting the correctness of the model’s output. This class of efficiency attacks bypasses traditional robustness evaluations, which focus primarily on misclassification or semantic corruption.

Existing research has begun to surface examples of such attacks across different types of DDLS architectures, including early-exit classifiers, autoregressive generators, and dynamic object detectors~\cite{rathnasuriya2025efficiency}. However, the current understanding remains fragmented. Prior work typically targets isolated architectures or behaviors, with limited attention to general principles or common failure modes~\cite{rathnasuriya2025efficiency}. Furthermore, modern model families, such as Mixture-of-Experts, gated transformers, and hierarchical pipelines have yet to be systematically examined under this threat model. As these architectures increasingly rely on dynamic computation to scale to real-world use cases, the absence of targeted defenses raises serious concerns about their resilience in adversarial settings.

In this work, we conduct a security-driven investigation into the efficiency vulnerabilities introduced by dynamic behaviors in DDLSs. We analyze how input-adaptive mechanisms, which are central to the efficiency gains of these systems, can also serve as adversarial entry points. We review existing efficiency attacks, assess their alignment with current architectural trends, and identify blind spots in attack coverage and mitigation strategies. Building on these insights, we outline a research agenda aimed at evaluating the feasibility of such attacks on emerging model classes and designing targeted defenses that preserve both computational efficiency and adversarial robustness.

This work initiates an investigation into the efficiency vulnerabilities of dynamic deep learning systems. Our contributions are as follows:
\begin{enumerate}
    \item We identify input-adaptive behaviors in DDLSs as a novel attack surface, enabling efficiency degradation without affecting output correctness.
    \item We provide a unified perspective on how existing attacks exploit these behaviors and highlight unaddressed risks in emerging architectures.
    \item We propose a forward-looking agenda to evaluate attack feasibility on modern DDLSs and develop defense mechanisms tailored to dynamic inference.
\end{enumerate}

\section{Background and Problem Statement}
\label{background}

\begin{figure*}[t]
    \centering
    
    \includegraphics[width=\linewidth]{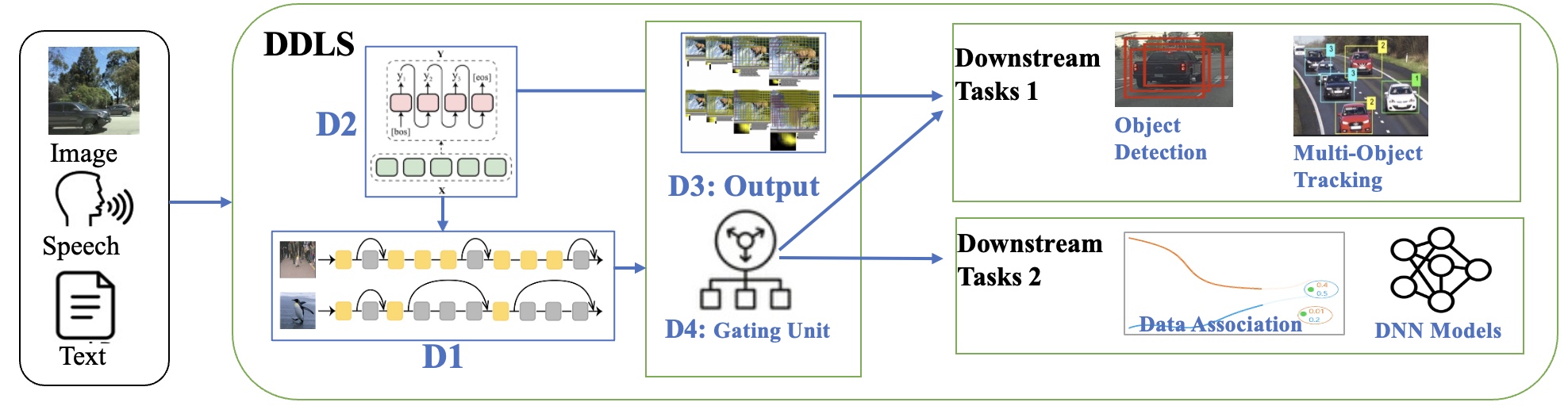}
   
    \caption{Illustration of Dynamic Behavior (\textbf{D}) on DDLS. \textbf{\texttt{D1}} examines the behavior where computational outputs fluctuate across individual inference iterations. \textbf{\texttt{D2}} focuses on attacks that alter the number of iterations needed to complete inference. \textbf{\texttt{D3}} involves attacks that escalate the number of generated outputs, thereby increasing the overall computational load. \textbf{\texttt{D4}} attacks that triggers gating mechanisms to force downstream modules to perform unnecessary high-cost computations.}
    \label{fig:ddls}
\end{figure*}

This section establishes the foundation for by providing a structured overview of efficiency attacks in DDLSs. We define the scope of efficiency attacks, categorize the dynamic behaviors inherent to DDLSs, and outline the relevant threat models.

\subsection{Definition of Efficiency Attacks}

Efficiency attacks exploit the adaptability of DDLSs to maximize computational cost. Unlike traditional adversarial attacks that induce misclassification, these attacks degrade system performance by inflating resource consumption. 

We define an efficiency attack as a constrained optimization problem. Given a neural network $\mathcal{N}(\cdot)$, an input $x \in \mathcal{X}$, and a hardware accelerator $\mathcal{H}$, the computational cost is represented as $\mathcal{C}(\mathcal{N}, \mathcal{H}, x)$ measured in energy consumption, latency, or FLOPs. The adversary seeks an optimal perturbation $\delta$ that maximizes computational cost while remaining imperceptible and within the valid input space:

\begin{equation}
\begin{split}
    &\text{maximize} \quad \mathcal{C}(\mathcal{N}, \mathcal{H}, x + \delta)\   \\ 
& \text{subject to} \quad ||\delta|| \leq \epsilon, \quad x + \delta \in \mathcal{X}
\end{split}
\end{equation}

The objective is to increase computational cost without violating input constraints ($||\delta|| \leq \epsilon$), exposing a fundamental vulnerability in efficiency-optimized DDLSs.

\subsection{Adversarial Exploitation of Dynamic Behaviors in DDLSs}

Figure~\ref{fig:ddls} depicts an overview of the progressive dynamic behaviors observed in DDLSs. When an input, such as an image, audio segment, or sentence, is fed into the system, different types of adaptive behavior may be triggered. At the finest level, \textbf{dynamic computation per inference} (\texttt{D1}) allows the system to adjust the depth or width of its execution path using techniques like early exits, layer skipping, or token pruning. At a coarser level, \textbf{dynamic inference iterations} (\texttt{D2}) emerge in autoregressive or iterative models, where the number of computational steps varies depending on convergence or generation length. Further, \textbf{dynamic output production} (\texttt{D3}) arises in tasks like object detection or segmentation, where the number of generated outputs (e.g., bounding boxes or masks) is input-dependent and affects downstream processing load. These three behaviors (\texttt{D1}–\texttt{D3}) were characterized in our prior work~\cite{rathnasuriya2025efficiency}.

In this work, we introduce and formalize a fourth behavior: \textbf{dynamic gating mechanisms} which operate at the system’s entry points. These modules decide whether an input should trigger heavy computation or be filtered early using lightweight classifiers, confidence thresholds, or decision heuristics. Gating is especially common in multi-stage pipelines, where only selected inputs are routed to expensive components.


While such dynamic behaviors improves efficiency by tailoring inference cost to input complexity, they collectively introduce execution variability that is externally observable and input-controllable. As a result, adversaries can exploit this variability to manipulate computational load, effectively converting a model’s adaptive behavior into an attack surface.

\subsection{Threat Model}
\label{threat}

The adversary’s objective is to degrade the efficiency of a DDLS by maximizing computational cost during inference. By introducing imperceptible perturbations or manipulating training data, the attacker prevents the system from leveraging its adaptive behavior, thereby inducing excessive resource consumption.

\noindent\textbf{Adversary Goals.} The adversary seeks to:
(i) maximize inference-time computational cost (e.g., latency, FLOPs, or energy),
(ii) ensure perturbations are imperceptible to humans or within application constraints, and
(iii) preserve real-world plausibility so that adversarial inputs remain valid in practice.

\noindent\textbf{Adversary Capabilities.} We consider two primary attack vectors: First, \textbf{evasion attacks} occur at inference time. In the \textit{white-box} setting, the adversary has full access to the model's architecture, parameters, and internal activations, enabling precise manipulation of inputs to disable dynamic behaviors.
In the \textit{black-box} setting, the adversary lacks internal access but can observe side-channel signals such as inference time or energy usage. Attacks are guided by iterative queries, surrogate modeling, or optimization techniques to infer dynamic behavior and exploit it. Second, \textbf{poisoning attacks} occur during training to embed inefficiencies that persist at inference. These attacks assume that the
adversary can inject training data or modify model updates. In \textbf{data poisoning}, adversaries craft inputs to bias model behavior toward increased computational cost. In \textbf{model poisoning}, the training procedure is manipulated to encode inefficient execution paths. Both approaches raise inference-time costs, undermining dynamic mechanisms designed for efficiency.

\section{Analysis and Preliminary Findings}
\label{study}


Prior work has demonstrated that efficiency attacks can exploit range of dynamic behaviors in DDLSs. However, existing techniques typically target isolated architectural patterns, such as early exits, output extension, or layer skipping, without generalizing to broader classes of dynamic systems. For example, attacks like DeepSloth~\cite{hong2020slowdownattack}, NICGSlowdown~\cite{NICGSlowDown}, and LLMEffiChecker~\cite{feng2024llmeffichecker} focus on particular mechanisms (e.g., \texttt{D1} or \texttt{D2}) but do not address the full space of modern DDLS behaviors, especially \texttt{D3} and \texttt{D4}~\cite{rathnasuriya2025efficiency}.

\begin{table}[ht]
\centering
\caption{The Average Effectiveness of LLMEffiChecker in Degrading LLaMA 3B Performance under Word and Character Attacks for White-box (W) and Black-box (B) Settings}
\label{tab:flops-latency-energy}
\resizebox{\linewidth}{!}{
\begin{tabular}{l|ccc|ccc|ccc}
\toprule
& \multicolumn{3}{c|}{\textbf{FLOPs}} & \multicolumn{3}{c|}{\textbf{Latency}} & \multicolumn{3}{c}{\textbf{Energy}} \\
\textbf{Attack Type} & $\epsilon{=}1$ & $\epsilon{=}2$ & $\epsilon{=}3$ & $\epsilon{=}1$ & $\epsilon{=}2$ & $\epsilon{=}3$ & $\epsilon{=}1$ & $\epsilon{=}2$ & $\epsilon{=}3$ \\
\midrule
Character (B) & 29.25 & 36.27 & 37.27 & 89.58 & 128.01 & 133.55 & 89.09 & 127.58 & 132.22 \\
Word (B)      & 46.00 & 46.80 & 41.52 & 154.71 & 153.93 & 144.40 & 154.04 & 153.53 & 144.07 \\
Character (W) & 21.13 & 21.13 & 21.13 & 97.32 & 97.32 & 97.32 & 97.91 & 97.91 & 97.91 \\
Word (W)      & 42.56 & 42.56 & 42.56 & 149.54 & 149.54 & 149.54 & 148.91 & 148.91 & 148.91 \\
\bottomrule
\end{tabular}

}

\end{table}

Our systematization of knowledge~\cite{rathnasuriya2025efficiency} reveals that most existing methods operate under white-box assumptions, leveraging access to internal gradients, activations, or model control flows. Only a limited number of black-box strategies exist, typically relying on surrogate models or energy estimators. These include approaches such as SpongeExamples and EREBA, which demonstrate high energy inflation (up to 2000\%) in constrained scenarios. However, they often depend on precomputed transfer sets or costly evolutionary search, limiting their applicability to real-world systems. Notably, general-purpose and scalable black-box attacks on dynamic behaviors remain largely underexplored.


To assess the feasibility of efficiency attacks on large-scale models, we conducted preliminary experiments on LLaMA 3B model using a machine translation task. We evaluated both white-box (W) and black-box (B) attack settings. We applied LLMEffiChecker~\cite{feng2024llmeffichecker} to generate perturbations that increase generation cost in terms of average increase in latency, energy, and FLOPs compared to the benign input. Our results in Table~\ref{tab:flops-latency-energy} show that: (1) Word-level attacks increase GPU latency by 154.71\% and FLOPs by up to 46.8\%, with no significant drop in BLEU score, (2) Black-box attacks using are comparably effective, especially at word-level granularity, and (3) Character-level perturbations are less effective but show increasing cost with larger $\epsilon$ values. 


To evaluate defenses, we tested both general-purpose and model-specific mitigation strategies~\cite{rathnasuriya2025efficiency}. Input validation based on hidden states or gradients achieved a detection accuracy of upto 87\%, showing strong potential for detecting efficiency attacks without compromising task accuracy. On the other hand, low-cost input transformations such as JPEG compression and spatial smoothing, while effective against image-based attacks (e.g., DeepSloth), significantly degraded output quality in text-based models like NICGSlowdown. For instance, adversarially extended captions averaging 62 tokens were truncated to 9–12 tokens using JPEG, resulting in severe BLEU score degradation. These findings highlight that while defenses like input validation show promise, most existing techniques are either architecture-specific or unsuitable for general-purpose deployment. More robust and scalable strategies are needed to protect DDLSs, particularly in black-box settings and large models. 




\section{Future Work}
In light of the emerging vulnerabilities in DDLSs, our future work will focus on expanding the empirical understanding of attack feasibility and advancing defense strategies. We plan to investigate efficiency attacks on modern architectures that increasingly rely on dynamic execution, including gated transformers and Mixture-of-Experts models. These systems present unique challenges due to their routing logic and modular structure, which may expose new efficiency-related attack surfaces. In parallel, we aim to design defense mechanisms that are tailored to specific dynamic behaviors while maintaining a balance between computational efficiency and robustness. This includes behavior-aware input validation, dynamic execution constraints, and lightweight perturbation detection methods. Finally, we will explore runtime adaptive strategies, such as energy-based monitoring and control-plane gating that can detect and mitigate adversarial behavior without disrupting performance on benign inputs. Together, these directions seek to build a foundation for secure deployment of DDLSs in adversarial environments.

\section*{Acknowledgments}
This work was partially supported by NSF grants NSF CCF-2146443 and Amazon Research Award, Fall 2024.

\bibliographystyle{IEEEtran}
\bibliography{main}
\end{document}